\documentclass[sigconf]{acmart}
\usepackage{mathtools}
\DeclarePairedDelimiter{\ceil}{\lceil}{\rceil}
\AtBeginDocument{%
  \providecommand\BibTeX{{%
    \normalfont B\kern-0.5em{\scshape i\kern-0.25em b}\kern-0.8em\TeX}}}

\setcopyright{acmcopyright}
\copyrightyear{2020}
\acmYear{2020}
\acmDOI{10.1145/1122445.1122456}

\acmConference[CVMP 2020]{the 16th ACM SIGGRAPH European Conference
Visual Media Production}{Dec.\ 17--18}{London, UK}
\acmYear{2020}
\copyrightyear{2020}
\setcopyright{none}


\begin{document}

\title{High Fidelity Interactive Video Segmentation  Using Tensor Decomposition, Boundary Loss, Convolutional Tessellations, and Context-Aware Skip Connections}
\title[High Fidelity Interactive Video Segmentation]{High Fidelity Interactive Video Segmentation  Using Tensor Decomposition, Boundary Loss, Convolutional Tessellations, and Context-Aware Skip Connections}

\author{Anthony D. Rhodes}
\affiliation{Intel Corporation}
\email{anthony.rhodes@intel.com}
\author{Manan Goel}
\affiliation{Intel Corporation}

\email{manan.goel@intel.com}

\begin{abstract}
We provide a high fidelity deep learning algorithm (HyperSeg) for interactive video segmentation tasks using a convolutional network with \textit{context-aware skip connections}, and compressed, ”hypercolumn” image features combined with a \textit{convolutional tessellation} procedure. In order to maintain high output fidelity, our model crucially processes and renders all image features in high resolution, without utilizing downsampling or pooling procedures. We maintain this consistent, high grade fidelity efficiently in our model chiefly through two means: (1) We use a statistically-principled, tensor decomposition procedure to modulate the number of hypercolumn features and (2) We render these features in their native resolution using a convolutional tessellation technique. For improved pixel-level segmentation results, we introduce a boundary loss function; for improved temporal coherence in video data, we include temporal image information in our model. Through experiments, we demonstrate the improved accuracy of our model against baseline models for interactive segmentation tasks using high resolution video data. We also introduce a benchmark video segmentation dataset, the \textit{VFX Segmentation Dataset}, which contains over 27,046 high resolution video frames, including green screen and various composited scenes with corresponding, hand-crafted, pixel-level segmentations. Our work presents an extension to improvement to state of the art segmentation fidelity with high resolution data and can be used across a broad range of application domains, including VFX pipelines and medical imaging disciplines.   

\end{abstract}

\begin{teaserfigure}
  \includegraphics[width=\textwidth]{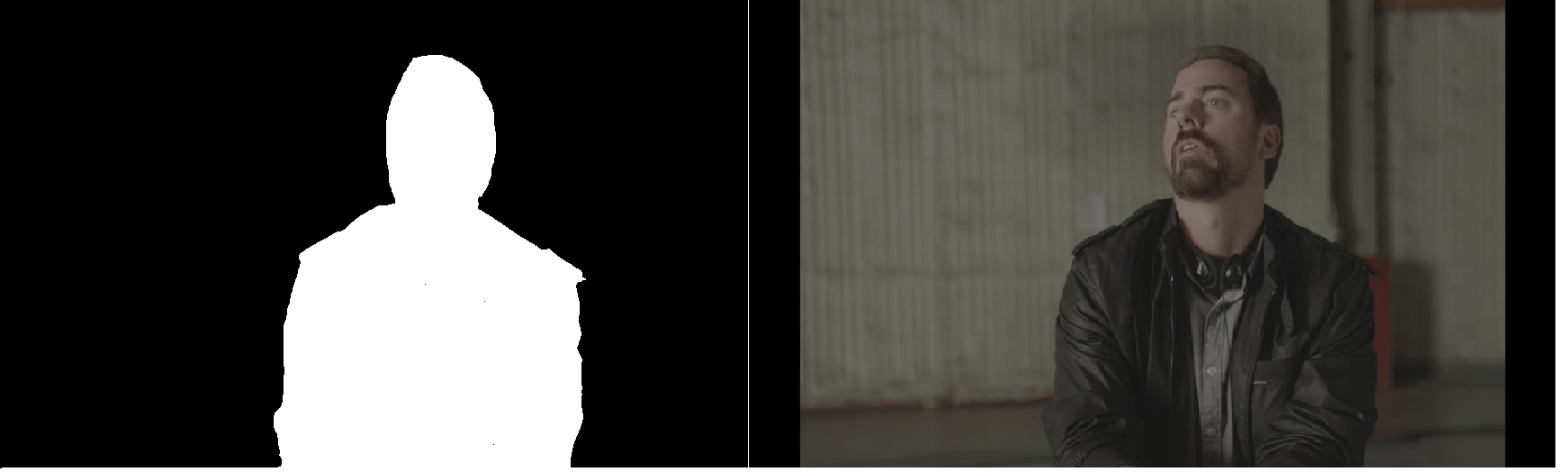}
  \caption{Rendered segmentation (left); input image, 2k resolution (right).}
    \label{fig:teaser}
\end{teaserfigure}




\maketitle

\section{Introduction}
\indent Image segmentation has a long history in computer vision and remains today an essential and largely unsolved problem. Prior to the advent of deep learning, classical methods for image segmentation included techniques such as thresholding \cite{journals/cvgip/LeeCP90},\cite{1176934}, clustering \cite{1455596},\cite{Chuang} compression \cite{10.1016/j.cviu.2007.07.005}, \cite{10.5555/104745}, historgram-based approaches \cite{Ni}, watershed \cite{Nguyen03}, region-growing \cite{Zhu1996}, in addition to techniques utilizing hand-crafted features, including \cite{Lowe04distinctiveimage}, \cite{335784}, \cite{6313729} -- to name a small number of methods. More recently, many algorithms employing energy minimization frameworks, including graph cuts \cite{Boykov01graphCuts}, Markov Random Fields \cite{10.1016/j.patcog.2004.04.015}, and related probabilistic graphic models \cite{10.1016/j.patcog.2004.04.015} have been implemented successfully. \\
\indent With the adoption of deep learning, convolutional neural networks (CNNs) \cite{Lecun98gradient-basedlearning}, \cite{10.5555/2999134.2999257} have become a standard model for data-driven approaches to computer vision tasks. Due to the early primacy of deep classification models across computer vision, many deep learning segmentation and localization algorithms rely on the appropriation of pre-trained classifiers as a key ingredient in their workflows. In particular, so-called region-proposal and related regression-classification methods \cite{zhao2018object}, including R-CNN \cite{10.1109/CVPR.2014.81}, Fast R-CNN \cite{10.1109/ICCV.2015.169}, Faster R-CNN \cite{10.5555/2969239.2969250}, FPN \cite{Lin}, and Yolo \cite{yolo}, all generate approximate, coarse-level object detections (i.e. bounding-boxes) for a set of class-specific objects. More recently still, deep learning classification pipelines have inspired methods in semantic segmentation tasks (i.e. pixel-level segmentation for class-specific objects) using "fully convolutional" networks; these approaches include, notably, FCN \cite{fcn}, Mask-RCNN \cite{He2017MaskR}, U-Net \cite{RFB15a}, and Deeplab (v2, v3) \cite{journals/corr/ChenPK0Y16}. \\
\indent In recent years, interactive  video  and interactive image segmentation  have  emerged as key  components of advanced image editing, graphics effects and medical imaging applications \cite{8270673}, \cite{10.1145/1015706.1015719}, \cite{b9007df6e5aa4e319679f68c0c0f9739}. Segmentation problems, including rotoscoping for vfx workflows \cite{10.1016/j.cviu.2013.10.013} and bio-medical image annotation \cite{Zhou_2019_CVPR} represent essential, labor-intensive image-processing tasks today which rely crucially on human-computer interaction.\\
\indent  Fully automated methods for interactive segmentation present a significant challenge because of the inherent ambiguity, multi-modality and complexity underlying pixel-level segmentation tasks, making these approaches largely infeasible. On the other hand, interaction-intensive methods conversely run the risk of requiring too much time, making them impractical in general. For these reasons, most researchers follow a semi-supervised approach to segmentation problems. In this setting, user interaction is commonly instantiated in the form of a set of user clicks, typically consisting of both \textit{positive} clicks (i.e. clicks within the object of interest/foreground) and \textit{negative} clicks (i.e. clicks indicating background pixels); alternatively, strokes, splines or partial segmentations are also often used as modalities of user interaction \cite{aspline}, \cite{Boykov01graphCuts}, \cite{14656282}. Video segmentation is often more difficult than image segmentation due the occurrence of motion blur, occlusion, and the temporal coherence requirements of consecutive segmentations \cite{benard}.\\
\indent In contrast to semantic segmentation tasks for class-specific objects, interactive segmentation represents a class-agnostic, few-shot active learning paradigm \cite{Li2018InteractiveIS}. Seen in this way, the segmentation model is tasked with determining the appropriate pixel classification (i.e. foreground/background) based on a relatively small number of labeled examples provided by the user. Fully-convolutional networks (FCN), trained end-to-end on input images and segmentation masks, have been shown to successfully synthesize low-level features (e.g. textures, color information) with high-level features such as objectness and semantic information robustly for few-shot vision tasks, including interactive segmentation \cite{fcn}, \cite{Li2018InteractiveIS}. Several approaches rely on improving this synthesis by applying a post-processing technique following an initial DNN approximate segmentation \cite{journals/corr/ChenPK0Y16}; \cite{Chen2017PhotographicIS} combine FCNs with a CRF for boundary refinement, while \cite{10.1109/ICCV.2015.179} combine FCNs with RNNs as approximate mean-field inference. \\
\indent Our work presents a single, high capacity FCN for end-to-end interactive video segmentation, building significantly on \cite{Li2018InteractiveIS}. This research draws upon several related works, including in particular: DOS (Deep Interactive Object Segmentation) \cite{Xu2016DeepIO}  which learns high-level objectness using a FCN with features generated by applying a distance transform to user clicks; \cite{fcn} who employ a skip architecture in conjunction with an FCN to propagate semantic information for segmentation; and \cite{Li2018InteractiveIS}, which is most closely related to our work, and incorporates latent diversity into a single FCN to improve output segmentation quality. \\
\indent In total,  our research provides the following contributions: (1) We apply tensor decomposition to a set of dense convolutional features to achieve substantial model compression, while maintaining segmentation fidelity; (2) We define a novel boundary loss function to improve segmentation quality along the periphery of an object; (3) We introduce a convolutional tessellation technique to render pre-trained features in the native input resolution; (4) We design our network to incorporate "context-aware" skip connections, which propagate the input image and interactive features across the model, thereby avoiding the dilution of these features in the latter layers of the network; (5) Lastly, we present a new benchmark video segmentation dataset, the \textit{VFX Segmentation Dataset}, containing 27,046 high resolution video frames with pixel-level ground-truth segmentations for training and testing segmentation models.  \\
\indent In the following sections, we present the details of our algorithm, followed by a comparative evaluation with several baseline and state of the art models on high resolution data. Our experiments demonstrate that the present approach outperforms the comparative models with respect to both overall segmentation accuracy (IOU -- intersection over union) and boundary segmentation accuracy (BIOU). In addition, in contrast to the comparison models, our model is the only network that processes the input data in its native resolution.

\section{Overview}
Our objective is to design a network $f$ that receives as input consecutive color images $\mathbf{X}_{t-1},\mathbf{X}_{t} \in \mathbb{R}^{w \times h \times 3} $ from video data in addition to user click features, and produces a binary mask that appropriately segments the object of interest indicated by the user. Concretely,  if we denote the input representation of the network as $\mathbf{X'}$, our goal is to train $f$ so that $f(\mathbf{X')}=\mathbf{Y} \in [0,1]^{w \times h}$, where the final binary output segmentation is generated by thresholding $Y$ at $1/2$.

\begin{figure}
    \centering
    \includegraphics[width=0.50\textwidth]{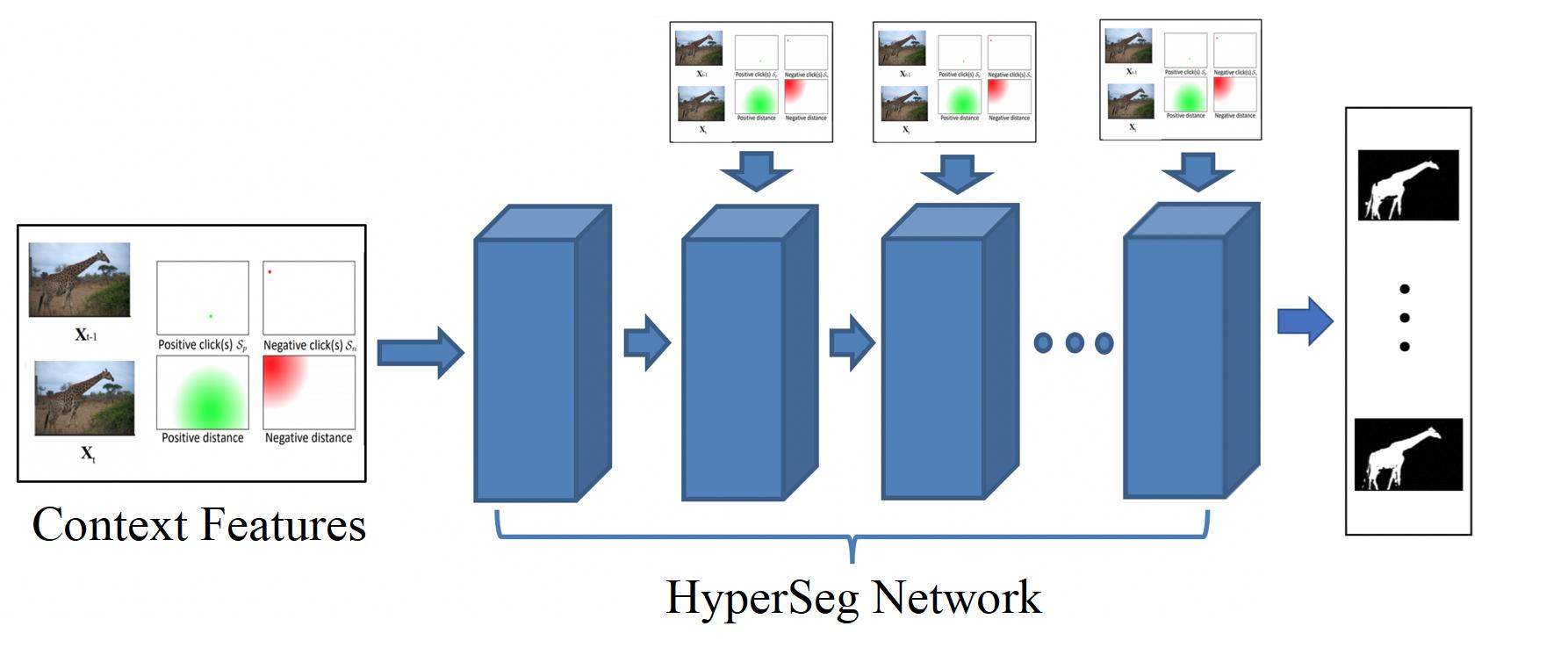}
    \caption{Segmentation network with context-aware skip connections schematic.}
    
    \label{fig:mesh1}
\end{figure}
\subsection{Segmentation Network}
\subsubsection{Network Architecture.} The input to our network $f$ includes the current and previous input frames, $\mathbf{X}_{t}$ and $\mathbf{X}_{t-1}$, and binary positive and negative user click masks $\mathbf{B}_{p}, \mathbf{B}_{n} \subset \{0,1\}^{w \times h}$. In addition, as in \cite{Li2018InteractiveIS}, we include diffusion maps $\mathbf{D}_{p}$ and $\mathbf{D}_{n}$ based on a distance transform applied to the binary user click masks; to this end, we define: $
\mathbf{D}_{n}(\mathbf{p})=\min_{\mathbf{q} \in S_{n}}{\| \mathbf{p}-\mathbf{q} \|}_{2}$ and $ \mathbf{D}_{p}(\mathbf{p})=\min_{\mathbf{q} \in S_{p}}{\| \mathbf{p}-\mathbf{q} \|}_{2}$.

We furthermore apply a VGG-19 network \cite{vgg} pretrained on the ImageNet dataset \cite{imagenet_cvpr09} to the
image  $\mathbf{X}_{t}$ as part of a convolutional  tessellation procedure, which we denote $T(\cdot)$ and detail in \textit{Section B}. These feature maps are then upsampled  to the resolution of $\textbf{X}_{t}$ using a \textit{nearest-neighbor} approximation. After generating these pre-trained features for the input image, we subsequently apply a tensor decomposition to these dense feature maps, denoted by the mapping $\phi(\cdot)$, and described in \textit{Section C}. In total, the final input to our network consists of the previous and current input frames $\mathbf{X}_{t-1} ,\mathbf{X}_{t}$, 
binary user click masks $\mathbf{B}_{p}, \mathbf{B}_{n}$, diffusion maps $\mathbf{D}_{p},\mathbf{D}_{n}$, and pre-trained features of the input image following the the convolution tesselation procedure and tensor decomposition algorithm, indicated by $T(\phi(\mathbf{X}_{t}))$. The input to the network consist of 746 total features at resolution $1920 \times 1080$ (736 pre-trained features and 10 "context" features, including $\mathbf{X}_{t-1} ,\mathbf{X}_{t},\mathbf{B}_{p}, \mathbf{B}_{n},\mathbf{D}_{p}$ and $\mathbf{D}_{n}$). Each subsequent layer in the network has output depth $70$ and input dimension $80$ ($70$ features from the previous layer output + $10$ features comprising the context-aware skip connections).\\
\indent The core architecture of the HyperSeg network is a context aggregation network (CAN) \cite{Chen2017FastIP}, \cite{Yu:2016:MCA}. In the first layer of the network, we apply an affine projection at full resolution (using $1 \times 1$ convolutions), where each per-pixel hypercolumn input feature is mapped to $\mathbb{R}^{70}$. In the subsequent layers of the network we apply  $3 \times 3$ convolutions with increasingly larger dilations at full resolution, followed by RELU activation \cite{nair2010rectified}. In addition, each layer after the first layer receives the aforementioned input "context" features via skip connections (see Table 1 and Section \textit{D}). As in \cite{Li2018InteractiveIS}, our network generates a diverse \textit{set} of \textit{M} potential segmentations, where the network is trained to exploit the inherent multimodality of plausible object segmentations extant in the model latent  space.  As \cite{Li2018InteractiveIS} show, exploiting this latent diversity results in significant improvements in pixel-level segmentation accuracy.
\begin{center}
\begin{table}
\begin{tabular}{ |c|c|c|c|c|c| } 
\hline
Layer & Convolution & Dilation & Pad & Input & Depth \\
\hline
1 & $1 \times 1$ & 1 & 0 & 746 & 70 \\ 
2 & $3 \times 3$ & 1 & 0 & 80 & 70 \\ 
3 & $3 \times 3$ & 2 & 2 & 80 & 70 \\ 
4 & $3 \times 3$ & 4 & 4 & 80 & 70 \\ 
5 & $3 \times 3$ & 8 & 8 & 80 & 70 \\ 
6 & $3 \times 3$ & 16 & 16 & 80 & 70 \\ 
7 & $3 \times 3$ & 32 & 32 & 80 & 70 \\ 
8 & $3 \times 3$ & 64 & 64 & 80 & 70 \\ 
9 & $3 \times 3$ & 128 & 128 & 80 & 70 \\ 
10 & $3 \times 3$ & 1 & 0 & 80 & 6 \\ 
\hline
\end{tabular}
\centering
\end{table}
\end{center}
\begin{figure}
    \centering
    \includegraphics[width=0.35\textwidth]{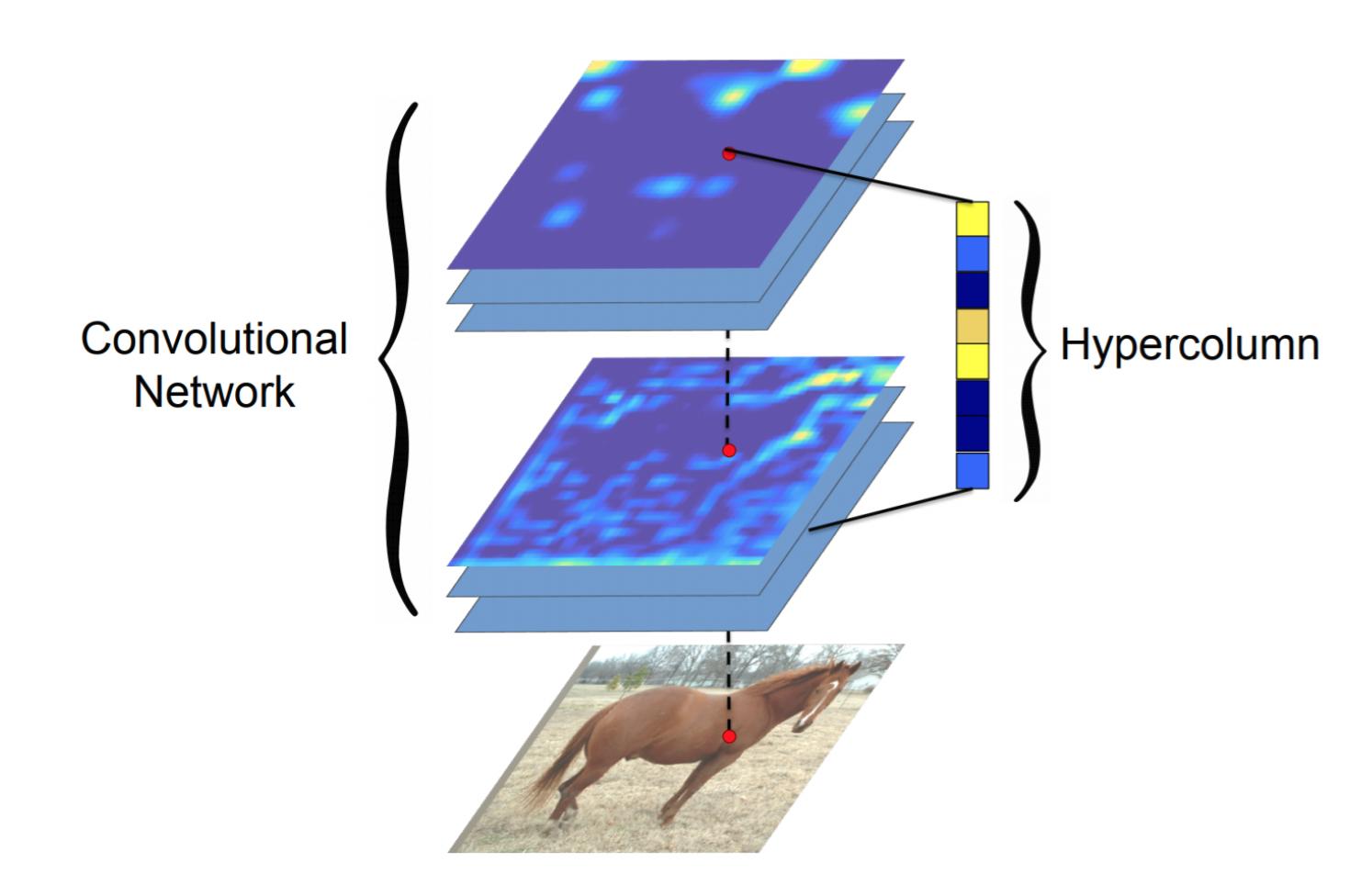}
    \caption{Example of dense, hyper-column image features generated by CNN. Image credit \cite{conf/cvpr/HariharanAGM15} }
    
    \label{fig:mesh1}
\end{figure}
\subsubsection{Loss Function}
We train the network $f$ on a loss function \cite{Li2018InteractiveIS} augmented with a novel component we term "boundary loss"; we define boundary loss:
\begin{equation}
\begin{gathered}
\sum_{i} \sum_{m=1}^{M}l_{\partial-PHL}(\mathbf{Y}_{i}, f_{m}(\mathbf{X}_{i};\mathbf{\theta}_{f}))
\end{gathered}
\end{equation}
where the outer sum is over the dataset, and the inner sum is over the \textit{M} segmentations produced by the network; $\mathbf{Y}_{i}$ denotes the ground truth segmentation of the \textit{i}th datum, and $f_{m}(\mathbf{X}_{i};\mathbf{\theta}_{f})$ indicates the \textit{m}th segmentation produced by the network with respect to the \textit{i}th datum. We use the subscript $l_{\partial-PHL}$ to stipulate use of the \textit{pseudo-Huber loss} \cite{barron2017general} applied to the \textit{boundary} (where $\partial$ denotes the boundary operator, i.e., we extract the largest peripheral contour of the given segmentation). The purpose of this additional loss function is to strongly encourage boundary points in the generated segmentation to match the ground-truth. \\
\indent We define the \textit{boundary pseudo-Huber loss function} as follows: 
\begin{equation}
\begin{gathered}
l_{\partial-PHL}(\mathbf{Y},f_{m}(\mathbf{X_{i}};\mathbf{\theta}_{f}))=\\
\sum_{\mathbf{p} \in \partial ( f_{m}(\mathbf{X_{i}};\theta_{f}))} \delta^{2} \Big( \sqrt{1+((Y(\mathbf{p})-f_{m}(\mathbf{X_{i}};\theta_{f})_{\mathbf{p}})/\delta)^{2}}-1 \Big)
\end{gathered}
\end{equation}
where the sum is performed over $\mathbf{p} \in \partial ( f_{m}(\mathbf{X_{i}};\theta_{f}))$, indicating the set of points along the boundary of the \textit{m}th segmentation generated by our network for the \textit{i}th datum. We use the notation $f_{m}(\mathbf{X_{i}};\theta_{f})_{\mathbf{p}}$ to indicate the value of the generated segmentation at the boundary point $\mathbf{p}$.  Pseudo-Huber loss combines the strengths of both $L_{1}$ and $L_{2}$ losses; the parameter $\delta$ controls the steepness of the loss function. In our experiments we use a relatively large $\delta$ value in order to severely penalize boundary segmentation errors. \\
\indent Using this definition of boundary loss, we define the following total loss function for our segmentation network:

\begin{equation}
\begin{split}
\mathscr{L}_{f}(\mathbf{\theta}_{f})=\sum_{i}\min_m \Big\{ l(\mathbf{Y}_{i},f_{m}(\mathbf{X}_{i};\mathbf{\theta}_{f}))+ l_{IC}(S_{p}^{i},S_{n}^{i},f_{m}(\mathbf{X_{i}};\mathbf{\theta}_{f}))\Big\} +\\
\sum_{i} \sum_{m=1}^{M}\lambda_{m}l(\mathbf{Y}_{i},f_{m}(\mathbf{X}_{i};\theta_{f}))+
\sum_{i} \sum_{m=1}^{M}l_{\partial-PHL}(\mathbf{Y}_{i},f_{m}(\mathbf{X_{i}};\theta_{f}))
\end{split}
\end{equation}
where above the first three terms of the loss function denote \textit{Jaccard loss}, \textit{interactive-context loss} and \textit{ranked diversity loss}, respectively, as defined in \cite{Li2018InteractiveIS}. \\
\indent Jaccard loss \cite{Berman2017TheLL} is a relaxation of IOU, defined for segmentation masks $A$ and $B$: 

\begin{equation}
l(A,B)=1-\frac{\sum_{\mathbf{p}}min(A(\mathbf{p}),B(\mathbf{p}))}
{\sum_{\mathbf{p}}max(A(\mathbf{p}),B(\mathbf{p}))}
\end{equation}
\indent In order to enforce agreement between the segmentation generated by our network and the user interaction provided via positive and negative clicks, we use an interactive-context loss function: 
\begin{equation}
\begin{gathered}
l_{IC}(S_{p}^{i},S_{n}^{i},A)= 
\|S_{p} \odot(S_{p}-A)\|_{1} + \|S_{n} \odot(S_{n}-(1-A))\|_{1}
\end{gathered}
\end{equation}
here  $\odot$ represents elementwise product. \\
\indent Lastly, \cite{Li2018InteractiveIS} demonstrate the utility of incorporating a ranked diversity loss function which imposes an ordering on the \textit{M} segmentations rendered by the network. This ordering is effectuated by multiplying each Jaccard loss term,  $l(\mathbf{Y}_{i},f_{m}(\mathbf{X}_{i};\theta_{f})),$ by a scalar value $\lambda_{m}$ from a decreasing sequence, which we define: $\lambda_{m}=10^{-2} \cdot 2^{M-m}$ for $1 \leq m \leq M.$ The effect of this ranked diversity loss function is to compel the model to break symmetries between solutions and to thereby consistently order the segmentations according to their accuracy (i.e. segmentation $m=1$ will in general produce the most accurate segmentation of the $M$ total segmentations produced by our network). 

\subsection{Dense Feature Compression}
Prior to passing the input image $\mathbf{X}_{t}$ through our segmentation network, we first generate image features from a pre-trained VGG-19 network. To do so, we pass $\mathbf{X}_{t}$ through the pre-trained network and extract feature maps from the following layers:  ‘conv1\_2’, ‘conv2\_2’, ‘conv3\_2’, ‘conv4\_2’, and
‘conv5\_2’ (denoted $C_{1},C_{2},C_{3},C_{4},C_{5}$, respectively). Next we upsample these features to the resolution of $\mathbf{X}_{t}$ using a \textit{nearest-neighbor} approximation, yielding a dense stack of hypercolumn features $H_{f} $ of dimension $1920 \times 1080 \times 1477$ \cite{conf/cvpr/HariharanAGM15}. We show a depiction of dense, hypercolumn features in Figure 3.Our network uses these dense, per-pixel VGG-19 features to account for the high degree of complexity inherent to pixel-level segmentation tasks. The downside, naturally, of using these dense features is their considerable memory cost. \\
\indent To reduce the memory and parameter overhead of our network, we apply a tensor decomposition to each of the tensors $C_{1},C_{2},C_{3},C_{4},C_{5}$. Concretely, we utilize a \textit{Higher Order Singular Value Decomposition} (HOSVD) \cite{conf/cvpr/HariharanAGM15},\cite{hosvd} algorithm known as \textit{Tucker decomposition} \cite{Kolda09tensordecompositions}, \cite{rabanser2017introduction} to specifically reduce the filter depth of each of these convolutional tensors. The goal of this compression is to diminish the overall depth of $H_{f}$ in a statistically-principled way whilst preserving the richness of the reduced VGG features inputted into our segmentation network; we refer to this compressed dimension as $D_{\phi}$. In our experiments we apply a roughly $2X$ compression rate using Tucker decomposition, yielding a compressed version of $H_{f}$ with dimension $1920 \times 1080 \times 736$ (i.e. $D_{\phi}=736)$.\\
\indent In general, for an \textit{N}-tensor, Tucker decomposition is framed as the problem of finding the decomposition of a tensor $\mathbf{X} \in \mathbb{R}^{R_{1} \times ... \times R_{N}}$, with $\mathbf{G} \in \mathbb{R}^{{R_{1}' \times ... \times R_{N}'}}$, and  $\mathbf{A}^{(1)} \in \mathbb{R}^{R_{1} \times R_{1}'}$,..., $\mathbf{A}^{(N)} \in \mathbb{R}^{R_{N} \times R_{N}'}$, where $1 \leq R_{i}' \leq R_{i}$ for $1 \leq i\leq N$,
 with the optimization constraint:
\begin{equation}
\begin{gathered}
\min_{\hat{\mathbf{X}}} \|\mathbf{X}-\hat{\mathbf{X}} \| \text{ with }
 \hat{\mathbf{X}} = \sum_{r_{1}=1}^{R_{1}} \cdot \cdot \cdot \sum_{r_{N}=1}^{R_{N}} g_{r_{1} \cdot \cdot \cdot  
r_{N}} \mathbf{a}_{r_{1}}^{(1)} \otimes \cdot \cdot \cdot \otimes \mathbf{a}_{r_{N}}^{(N)}\\
\end{gathered}
\end{equation}

where $\otimes$ connotes the tensor product; $\mathbf{G}$ is known as the \textit{core tensor} and the factor tensors $\mathbf{A}^{(1)},...,\mathbf{A}^{(N)}$ are called the \textit{principal components} of the decomposition. Compression is achieved when $R_{i}' < R_{i}$ for at least one axis of the tensor (note the strict inequality). In particular, because we wish to reduce the filter depth in each convolutional layer that we extract from VGG-19 while maintaining the original input image resolution, we apply the decomposition to each convolutional tensor along solely the depth axis. Figure 3 depicts a 3-tensor Tucker decomposition. 
\begin{figure}
    \centering
    \includegraphics[width=0.35\textwidth]{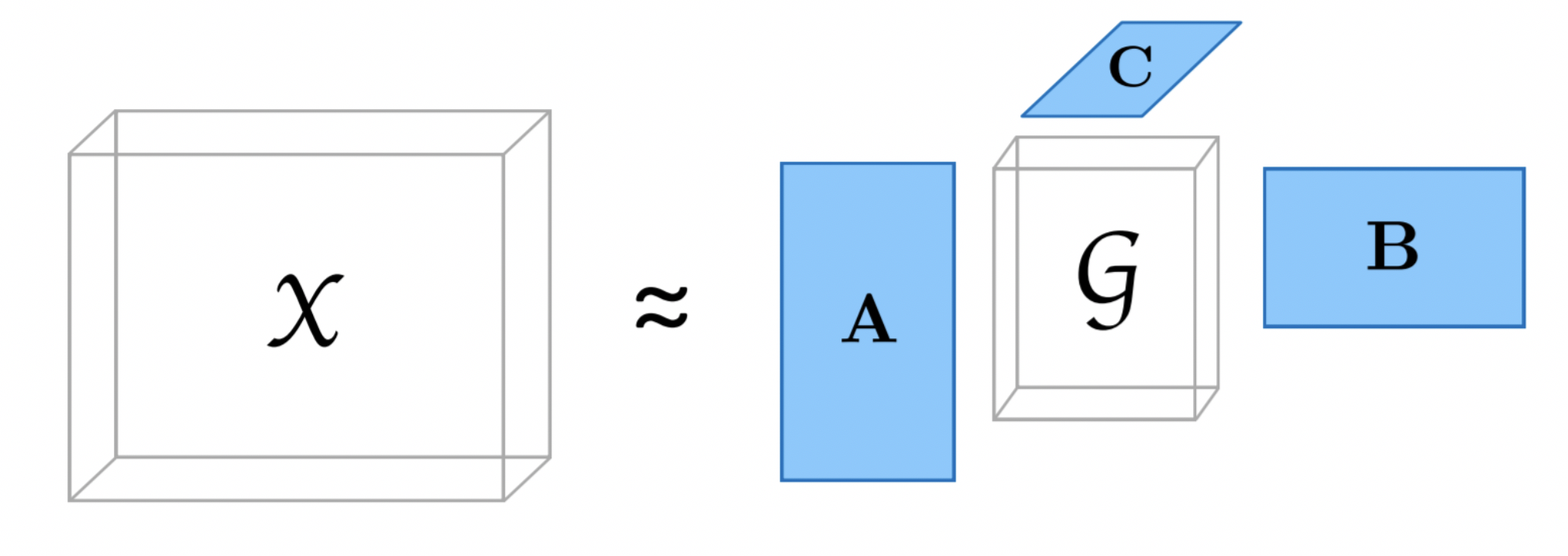}
    \caption{Depiction of 3-Tensor Tucker decomposition. Here $\mathbf{G}$ is the core tensor and $\mathbf{A}, \mathbf{B}, \text{ and } \mathbf{C}$ are the principal components of the decomposition of tensor $\mathbf{X}.$ Image credit \cite{rabanser2017introduction}.}
    \label{fig:mesh1}
\end{figure}
\subsection{Convolutional Tessellations}
Most standardized pre-trained deep models used in computer vision for feature extraction are classifiers (e.g., VGG \cite{vgg}, Resnet \cite{He2015DeepRL}, Inception \cite{43022}, etc.) trained on relatively low-resolution image data (ImageNet, for instance, has an average resolution of $462 \times 387$). The input resolutions of these standardized deep models are accordingly generally quite low -- for example, VGG-19 and Resnet both have input resolution $224 \times 224$). This low input image resolution can significantly impact the fidelity of the resultant features rendered by the deep network (see Figure 5). \\
\indent Pixel-level segmentation in high-resolution domains requires a very fine-grain output. As such, low-fidelity pre-trained features present a substantial barrier to improving deep learning models across a broad range of computer vision applications. We provide a simple and elegant solution to this problem using a convolutional tessellation algorithm. \\
\indent 
Given an input image $\mathbf{X}_{t}$ of dimension $w \times h$ and pre-trained model $M$ (e.g. VGG) with input resolution $w_{M} \times h_{M}$ we first interpolate $\mathbf{{X}_{t}}$ to dimension: $w_{m}\ceil{\frac{w}{w_{M}}} \times h_{m}  \ceil{\frac{h}{h_{M}}}$, where $\ceil{\cdot}$ is the ceiling function. The effect of this operation is to resize the input image to an integer multiple (by width and height, respectively) of the model width and height. This interpolation step resizes $\mathbf{X}_{t}$ into a grid of tiles, with $T=\ceil{\frac{w}{w_{M}}}\cdot \ceil{\frac{h}{h_{M}}}$ total tiles, each of dimension  $w_{m} \times h_{m}$. \\ 
\indent Following this interpolation step, we resize the image tensor of dimension $C \times w_{m}\ceil{\frac{w}{w_{M}}} \times$ $h_{m}\ceil{\frac{h}{h_{M}}}$ (where $C$ represents the number of color channels) into a $4D$ tensor consisting of stacked tiles of dimension $T \times C \times w_{M} \times h_{M}$; call this tensor $\mathbf{X}_{t}'$. Next, we pass $\mathbf{X}_{t}'$ through $M$, applying the previously described tensor decomposition procedure, yielding $\phi(M(\mathbf{X}_{t}')$. We process the tiles through $M$ as a minibatch of size $T$ so that this step can be parallelized for improved efficiency of our algorithm. \\
\indent After generating $\phi(M(\mathbf{X}_{t}'))$, a tensor of dimension $T \times D_{\phi} \times w_{M} \times h_{M},$ we finally construct a tessellation of these tile features with respect to their original location in $\mathbf{X}_{t}$ (before stacking), giving a tensor of size $C \times w_{m}\ceil{\frac{w}{w_{M}}} \times h_{m}\ceil{\frac{h}{h_{M}}}$. We follow this step with a final interpolation to render the tessellation in dimensions equal to the original image. Figure 4 provides a schematic of the deep convolutional tessellation algorithm; Figure 5 compares the fidelity of pre-trained VGG-19 features with and without our tessellation algorithm.  
\begin{figure}
    \centering
    \includegraphics[width=0.45\textwidth]{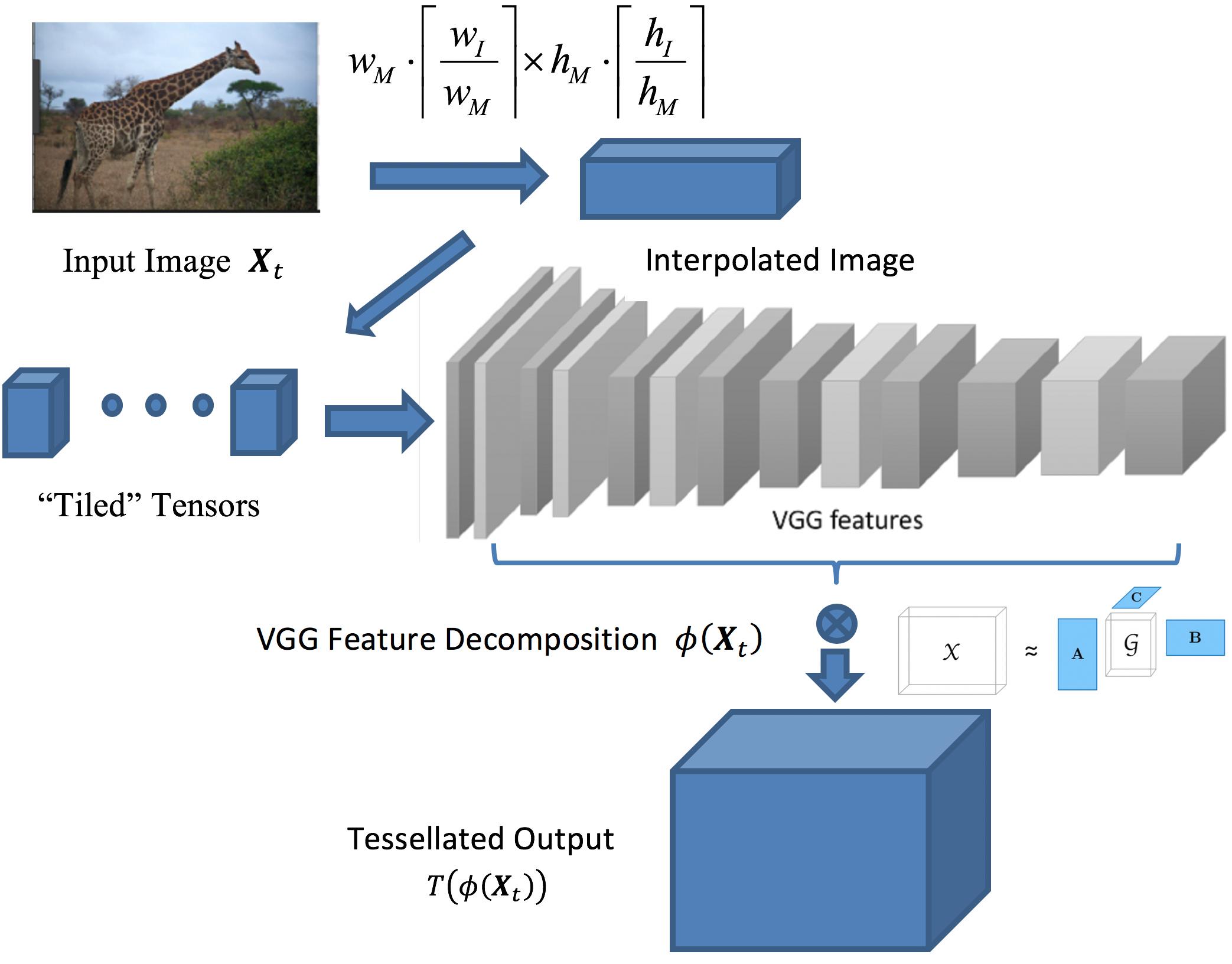}
    \caption{Schematic of the convolutional tessellation procedure.}
    
    \label{fig:mesh1}
\end{figure}

\begin{figure}
    \centering
     \includegraphics[width=0.33\textwidth]{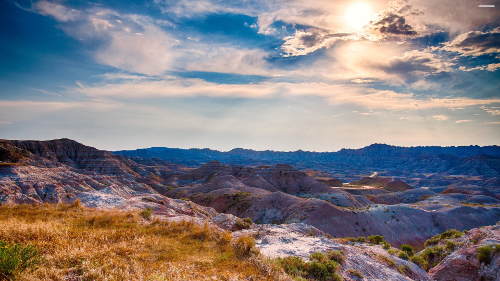} 
     
     \vspace{.5em}
     
    \includegraphics[width=0.33\textwidth]{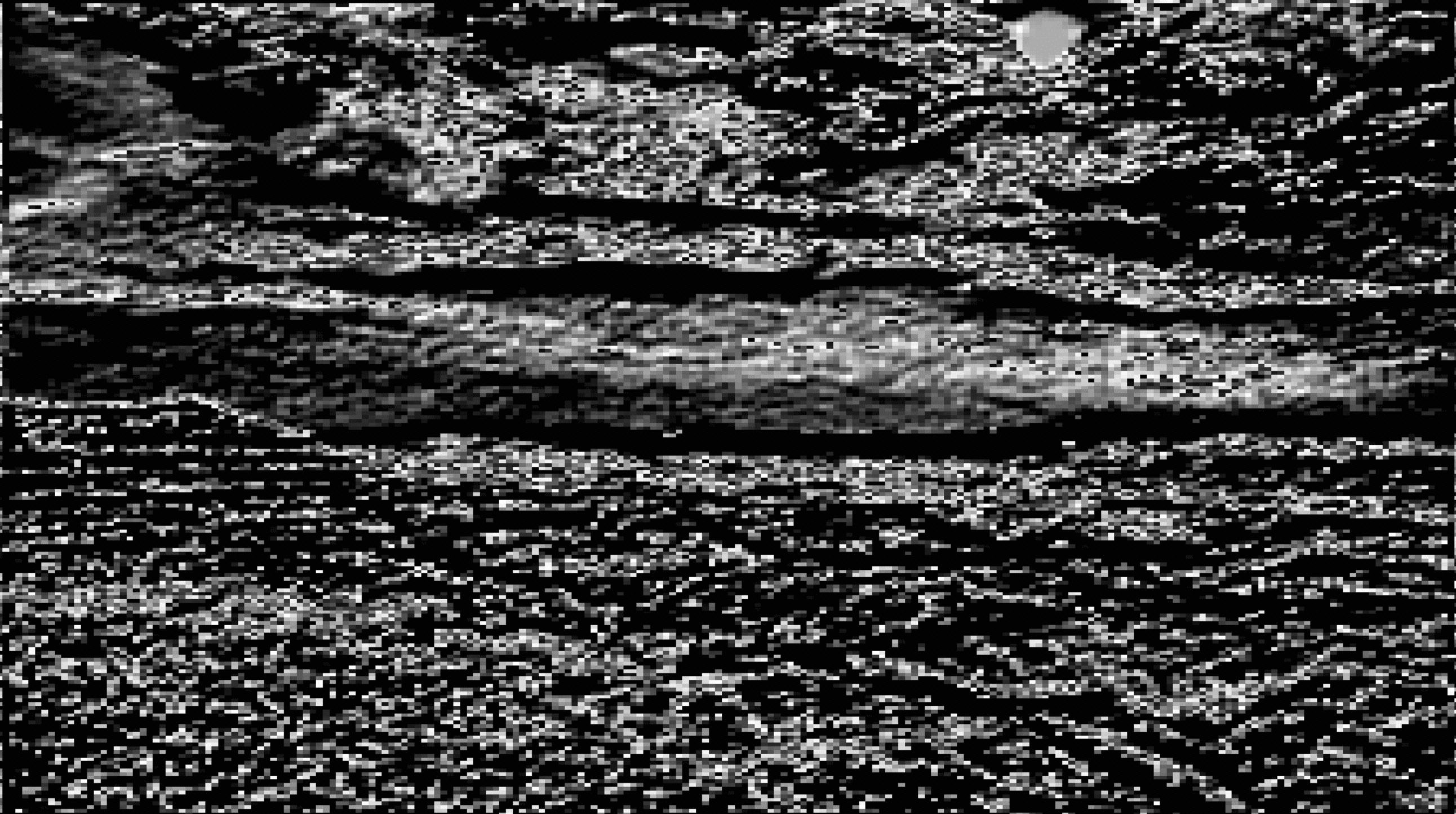}
    
    \vspace{.5em}
    
    \includegraphics[width=0.33\textwidth]{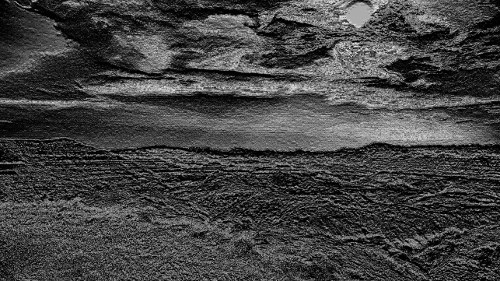}
    \caption{(Top) Original 2k resolution image; (Middle) non-tessellated deep convolutional feature map generated by VGG-19; (Bottom)  tessellated deep convolutional feature map using VGG-19.}
\end{figure}
\subsection{Context-Aware Skip Connections}
Recent research \cite{He2015DeepRL}, \cite{park2019SPADE} has amply demonstrated the value of adding so-called skip connections to deep architectures. Skip connections have been shown to mitigate phenomena such as \textit{vanishing gradient} and per-layer network saturation in deep learning. In general, skip connections are network wirings that allow features from early layers in the network to bypass later layers so that these features remain undiluted. These early features are typically (though not always) concatenated with activations in subsequent layers \cite{He2015DeepRL}. The basic intuition behind this approach is that the network is given access to a larger variety of feature representations (e.g. middle layer outputs), as opposed to a strict, hierarchically sequential representation of features. \\
\indent 
We introduce a novel variant of these concatenated skip connections in our segmentation network. Instead of traditional skip connections as described in the previous paragraph, we strictly propagate "context features" as skip connections across every layer in our network. By context features, we include the current frame image, the previous frame image, positive and negative user clicks, and positive and negative diffusion user clicks masks. Together, each of these features is concatenated with the previous layer output and then passed to the current layer of the network. \\
\indent We choose to pass context features in place of previous layer outputs for two basic reasons: (1) context features are high-fidelity, information-rich features with regard to the semantic segmentation task, and they are consequently highly \textit{discriminative}; (2) in addition, by propagating context features as skip connections, we are essentially providing the network with access to features that directly correlate with our model task; in this way the network does not need to reserve additional overhead to redundatly encode these task-relevant features in subsequent layers of the network. We show a schematic of context aware skip connections in Figure 2. 

\subsection{VFX Segmentation Dataset}
In addition to our segmentation network, we introduce an original, high-resolution dataset intended for pixel-level segmentation tasks in computer vision, the \textit{VFX Segmentation Dataset}. This dataset conists of 27,046 RGB video frames across $208$ different video clips, with each frame in $2k$ resolution ($1920 \times 1080).$ Our dataset consists  of professionally-filmed human subjects in studio, green screen, and composited scenes; each clip ranges in length from approximately 100-400 frames each. For each video frame, the ground-truth segmentation consists of either a keyed alpha mask (for green screen and composited videos) or a hand-rendered, binary pixel-level segmentation (for non-green screen content). For comparison, the current standard benchmark dataset for high-resolution video segmentation, DAVIS \cite{Caelles_arXiv_2019}, consists of $10,474$ frames across $150$ video sequences. The \textit{VFX Segmentation Dataset} is particularly suited for training and testing high fidelity segmentation, human subject tracking, and fostering robustness with composited video data (a common usecase with VFX studios). To maintain training parity with other segmentation models, we train our model on only a subset of this dataset in addition to the aforementioned DAVIS dataset (see Section III for details). We show several representative examples of video frames and their corresponding ground-truth segmentations from the \textit{VFX Segmentation Dataset} in Figure 7. 
\begin{figure}%
\centering
   \includegraphics[width=1.6in]{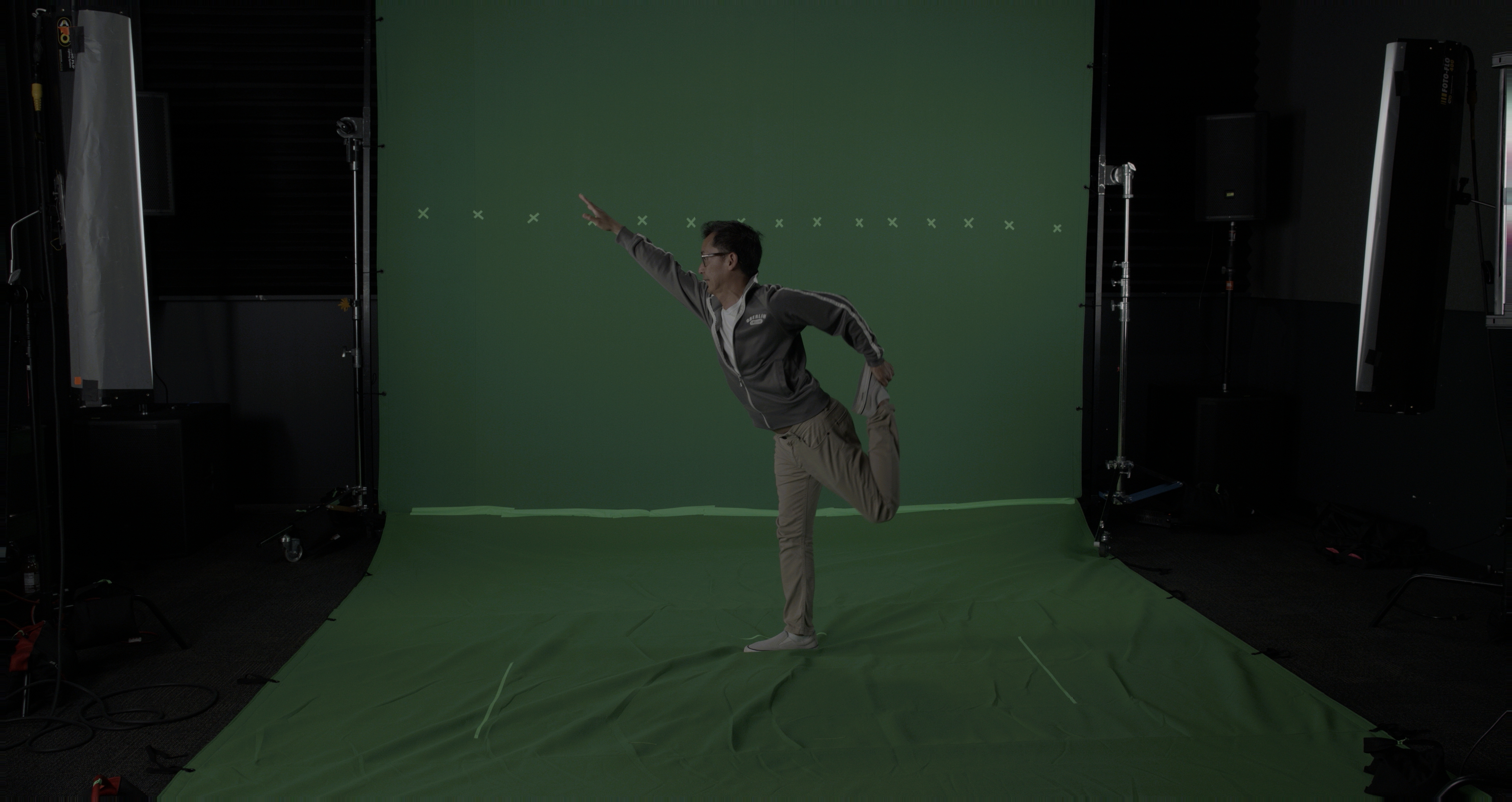}%
   \vspace{.5em}
  \includegraphics[width=1.6in]{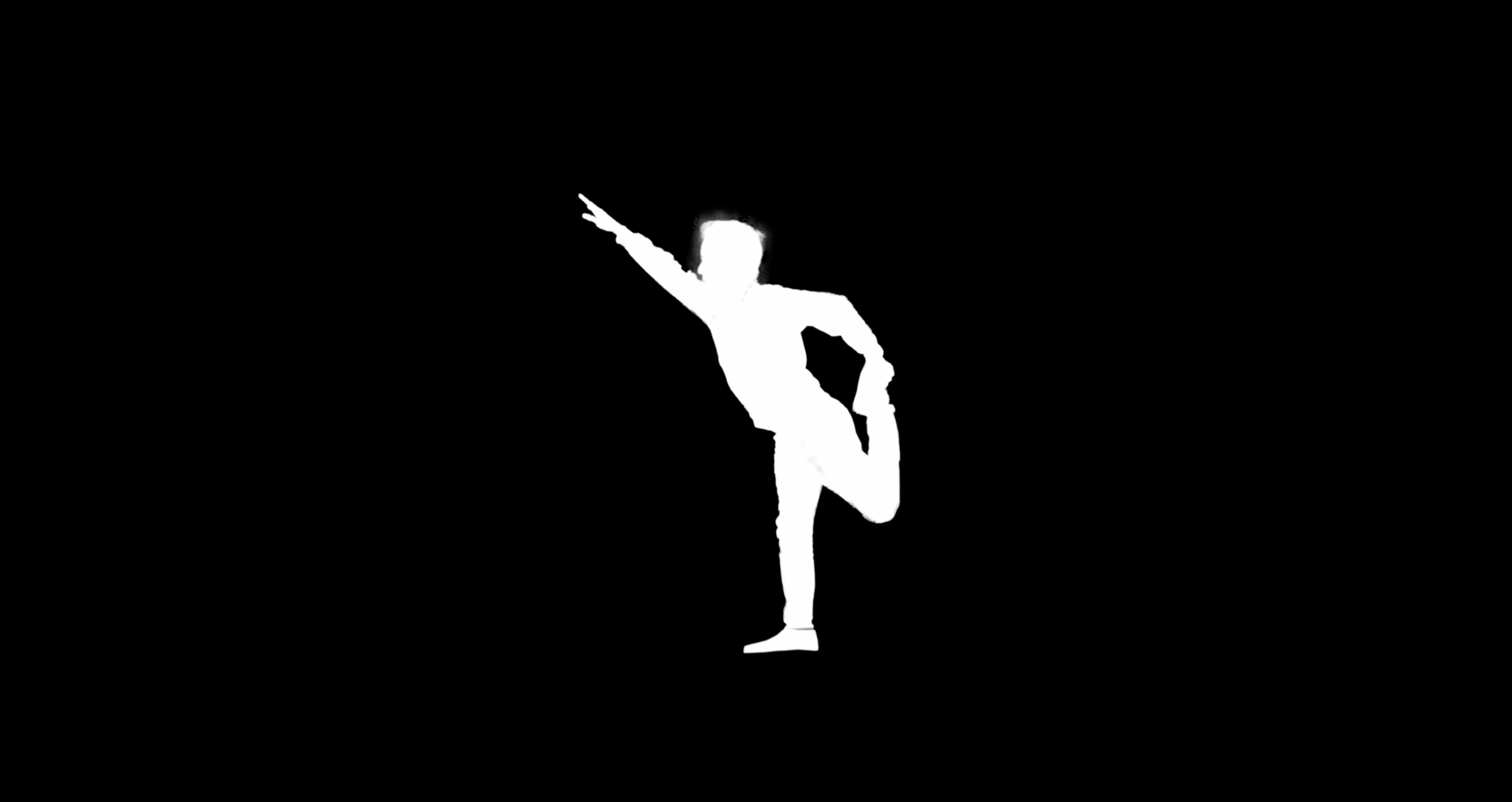}\\
  \includegraphics[width=1.6in]{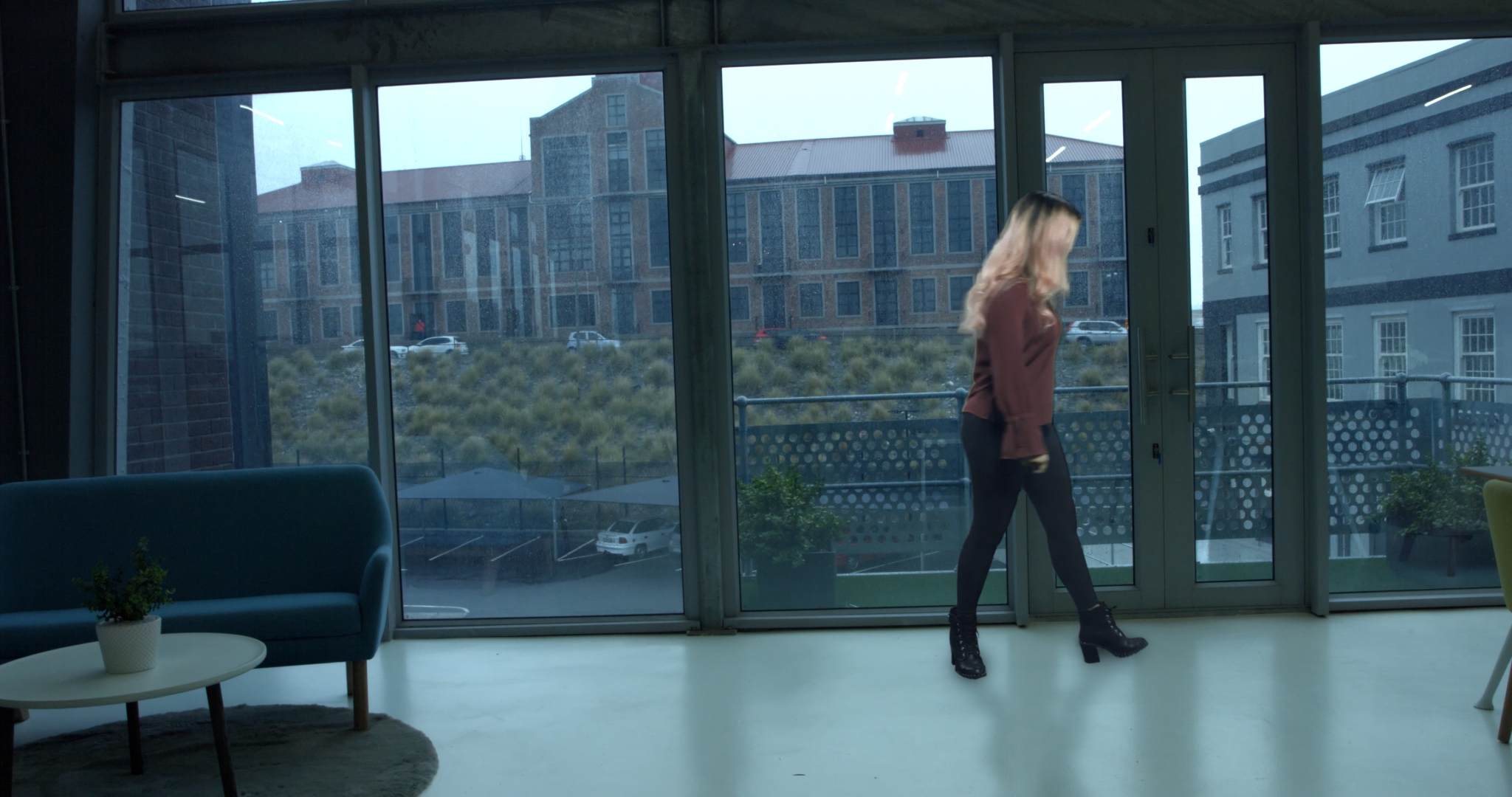}
  \includegraphics[width=1.6in]{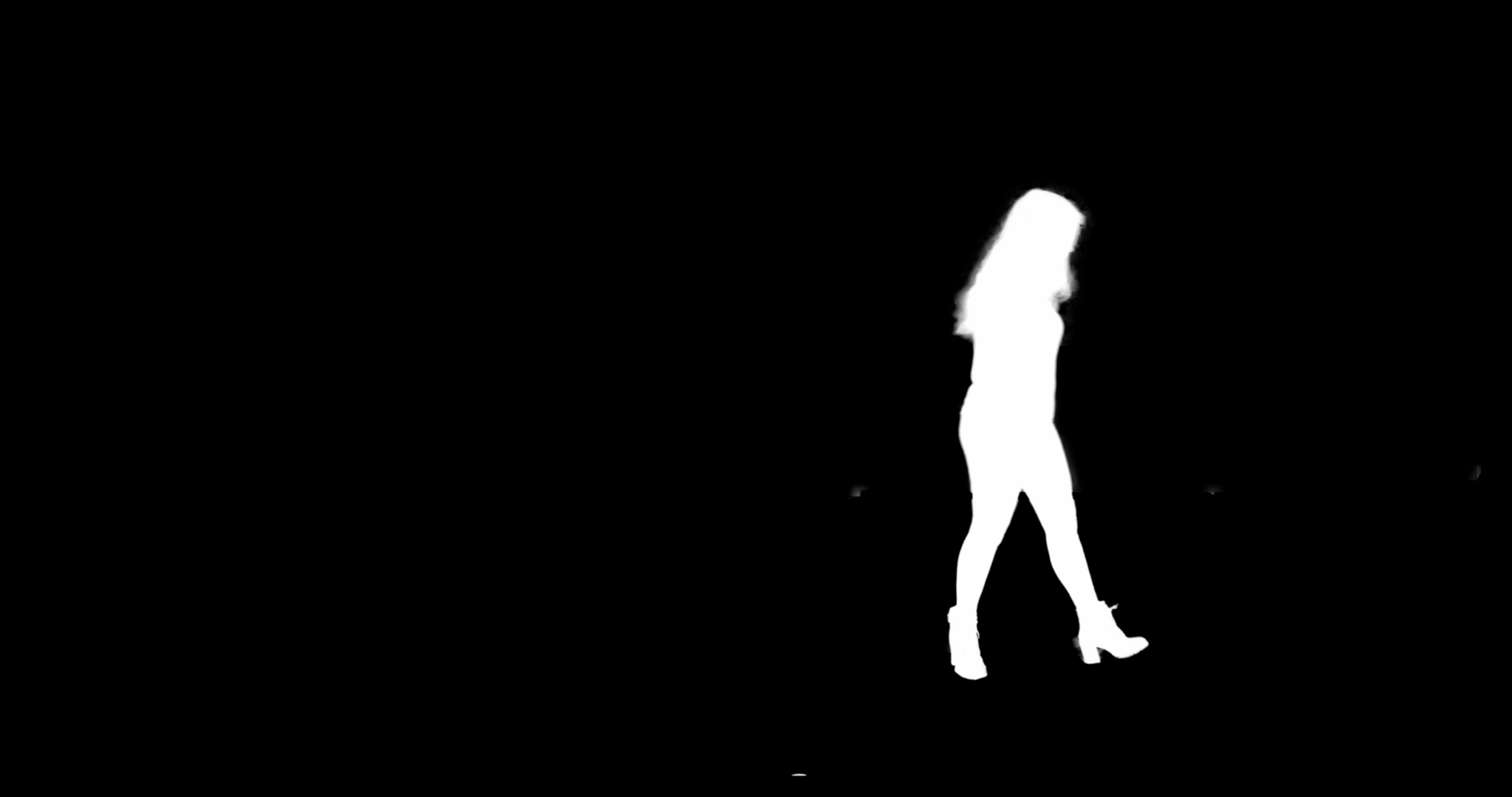}
  \caption{Representative images from the VFX Segmentation Dataset. (Left) Original RGB image; (Right) Hand-generated ground-truth segmentation.}

\end{figure}
\section{EXPERIMENTAL RESULTS}
Through experiments, we compare the segmentation results generated by our HyperSeg network with several state of the art and baseline models for the interactive image segmentation task. We use two metrics for this evaluation: mean IOU (mIOU) and mean boundary IOU (mBIOU). We define mBIOU as IOU applied between the boundary contours of the ground-truth and generated segmentation masks (after applying a dilation of 1 pixel to each segmentation boundary, respectively). Note, in particular, that mBIOU represents a very high standard for segmentation accuracy, as it reflects a near pixel-level match with the ground-truth segmentation boundary. HyperSeg was trained using 14,326 2k resolution images (8,761 images from the DAVIS dataset and + 5,565 from the VFX Segmentation Dataset) for 85 epochs. During training we generate simulated clicks (per image a random number of between 1-15 positive and negative clicks is rendered) using the click simulation strategy adopted in \cite{Xu2016DeepIO}. For testing, we used a randomly selected set of 1,622 2k resolution images ($1980\times 1020$) consisting of 849 images from the VFX Segmentation Dataset and 773 images taken from the DAVIS dataset; all the test images were held out from training. For evaluation, we fixed the interactive parameter using 10 total clicks (5 positive and 5 negative). In Figures 8 and 9 we show examples of representative HyperSeg segmentation results on the test data.\\
\indent In Table I we report experimental results for mIOU and mBIOU for HyperSeg compared with state of the art and baseline models, includingISEG (interactive segmentation) \cite{Li2018InteractiveIS}, DOS \cite{Xu2016DeepIO}, Graph Cut \cite{Boykov01graphCuts} and Random Walk \cite{10.1109/TPAMI.2006.233}. HyperSeg demonstrates a 19\% relative increase in mIOU and 34\% relative increase in mBIOU over SOA interactive segmentation models. In Figure 10 we show comparative examples of segmentation results with SOA models; in addition, Figures 11 and 12 highlight the improved smoothness and boundary accuracy exhibited by HyperSeg segmentations; Figure 11 additionally indicates the efficacy of our use of convolutional tensor decomposition for the HyperSeg segmentation output. \\

\begin{table}
\begin{center}

\centering
\caption{Segmentation Results}
\begin{tabular}{ |c|c|c|c| } 
\hline
Method& mIOU&mBIOU\\
 \hline
 \textbf{HyperSeg} (Ours)   &\textbf{ 0.840}    &\textbf{0.097}\\
 ISEG \cite{Li2018InteractiveIS}  & 0.705    &0.072\\
 DOS \cite{Xu2016DeepIO} & 0.681    &0.054\\
 Graph Cut \cite{Boykov01graphCuts} & 0.563    &0.0\\
 Random Walk \cite{10.1109/TPAMI.2006.233} & 0.639    &0.021\\
 \hline
\end{tabular}

\centering

\end{center}
\end{table}

\begin{figure}
    \centering
    \includegraphics[width=0.50\textwidth]{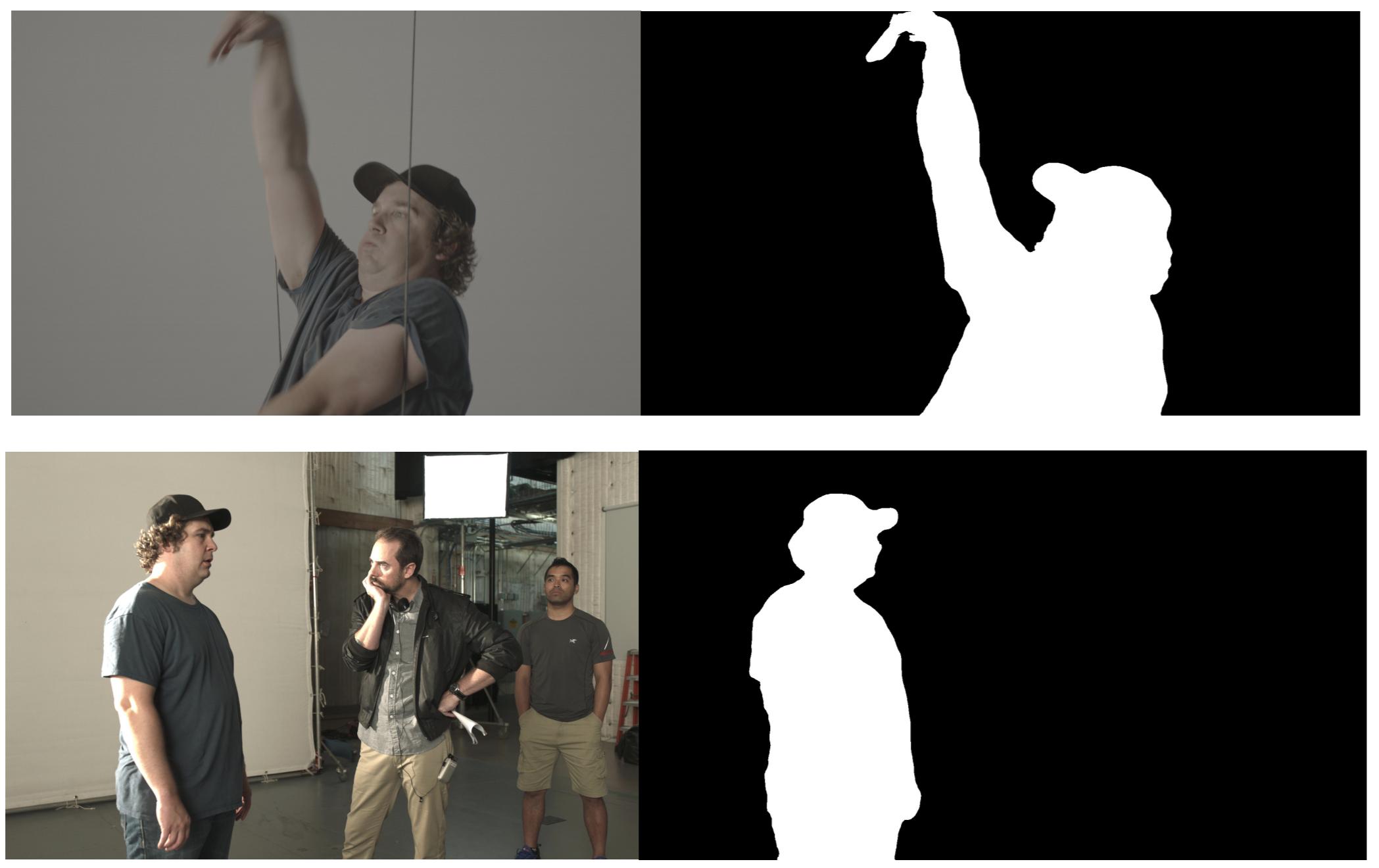}
    \caption{Example HyperSeg segmentation outputs from test data taken from the VFX Segmentation dataset (10 total clicks).}
    
    \label{fig:mesh1}
\end{figure}

\begin{figure}
    \centering
    \includegraphics[width=0.50\textwidth]{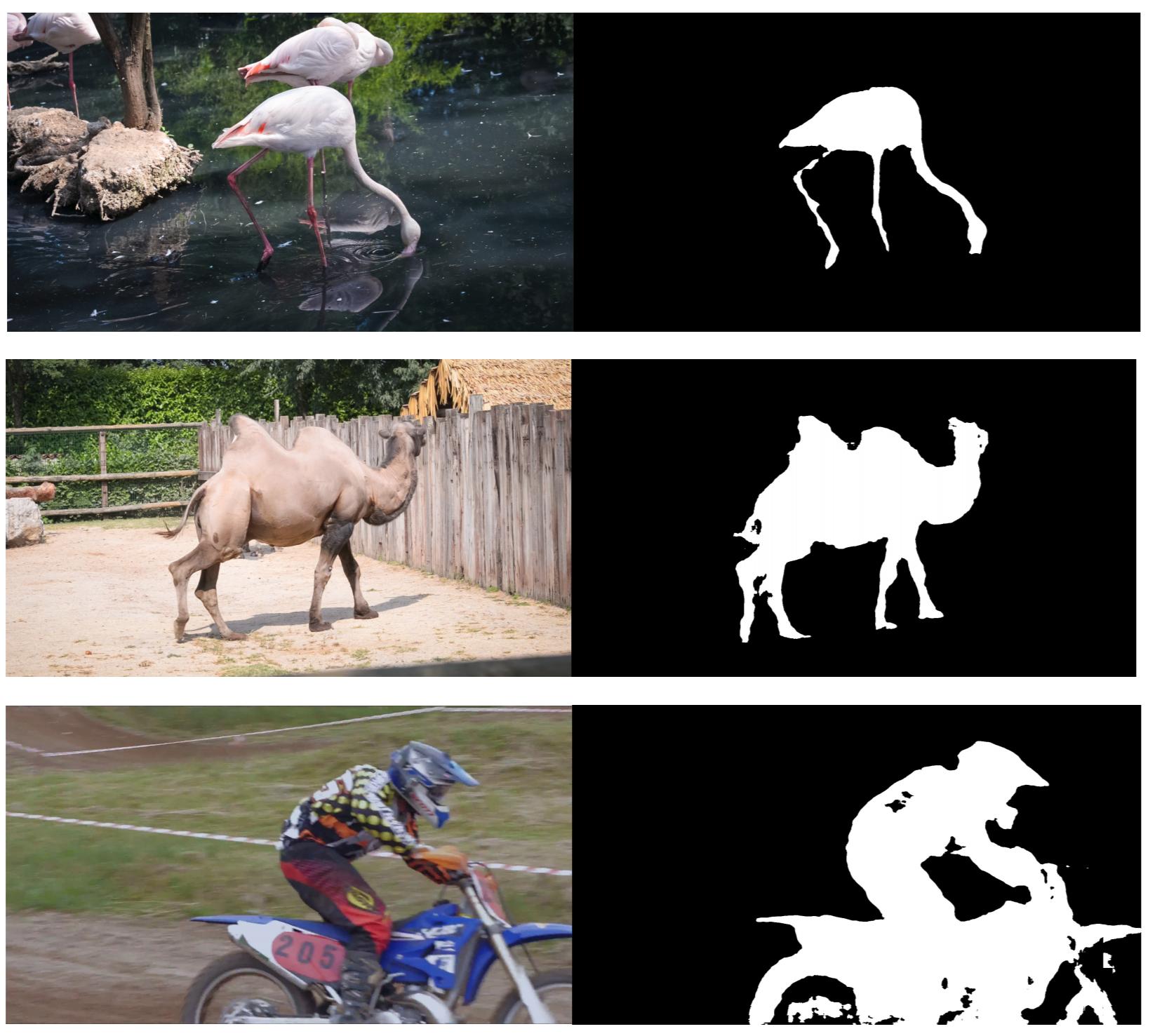}
    \caption{Example HyperSeg segmentation outputs from subset of test data taken from the DAVIS dataset (10 total clicks).}
    
    \label{fig:mesh1}
\end{figure}

\begin{figure}
    \centering
    \includegraphics[width=0.50\textwidth]{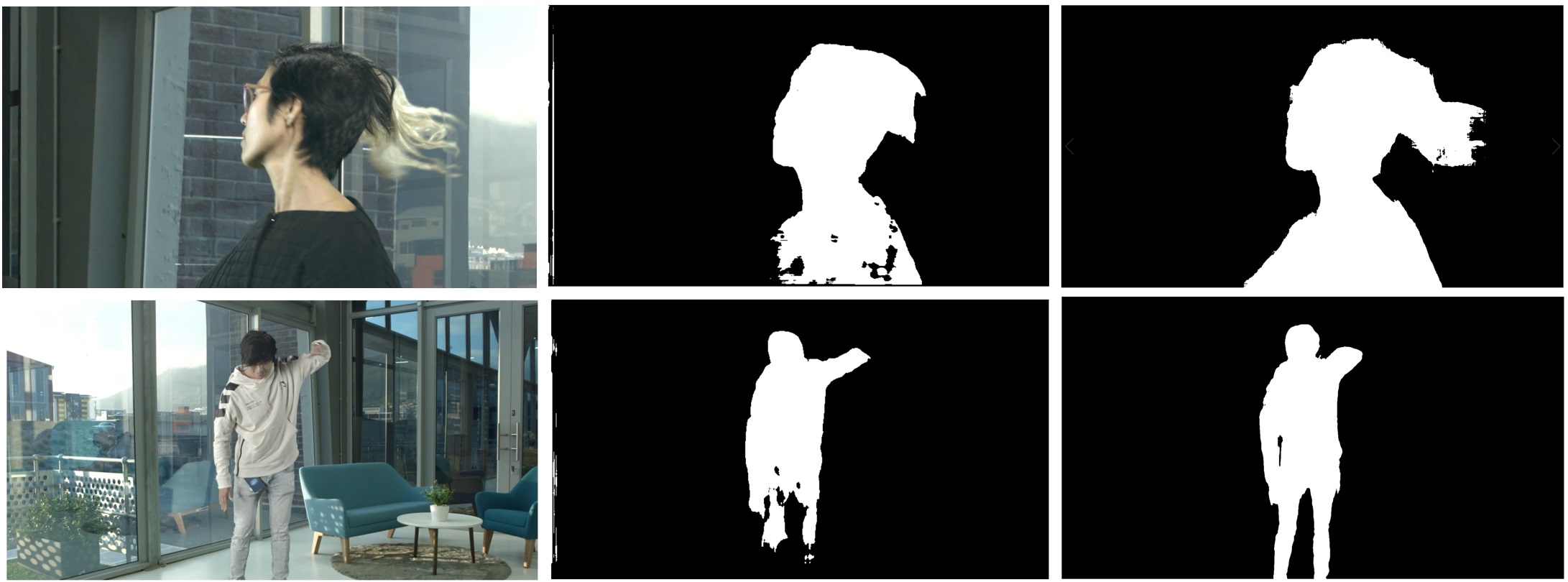}
    \caption{Segmentation quality comparison: (Left) Original 2k input image, (Middle) ISEG \cite{Li2018InteractiveIS}, (Right) HyperSeg (10 total clicks).}
    
    \label{fig:mesh1}
\end{figure}

\begin{figure}
    \centering
    \includegraphics[width=0.44\textwidth]{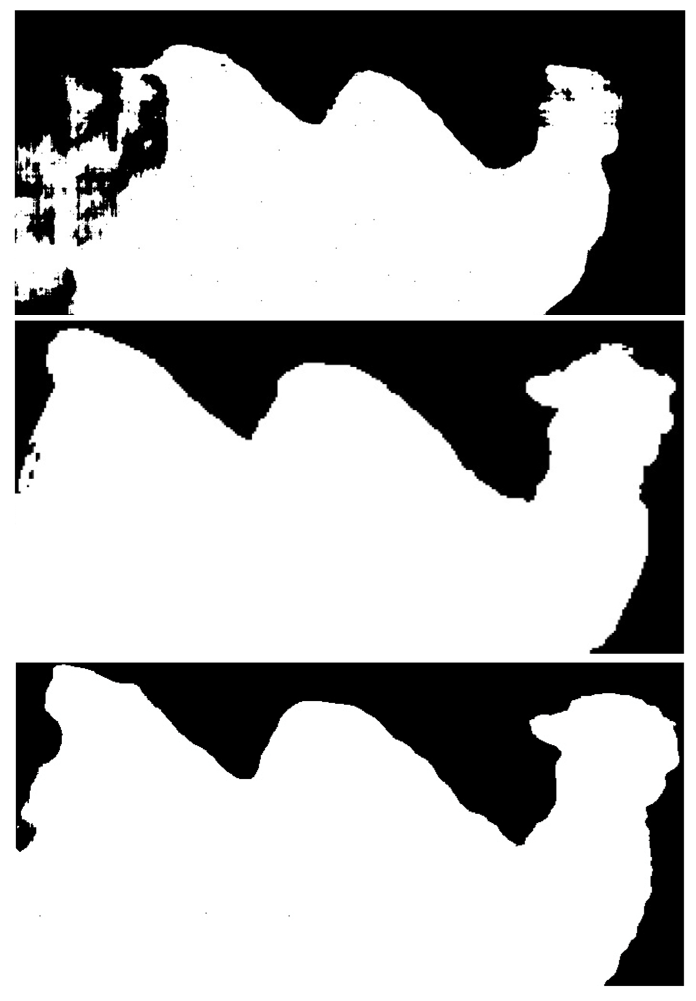}
    \caption{Tensor decomposition and segmentation boundary quality comparison: (Top) result for HyperSeg network trained without Tucker decomposition, using instead a random subset of $D_{\phi}=736$ dense, pre-trained VGG features; (Middle) ISEG \cite{Li2018InteractiveIS}, (Bottom) HyperSeg with Tucker decomposition; HyperSeg renders smoother, more accurate boundaries in general.}
    
    \label{fig:mesh1}
\end{figure}
\begin{figure}
    \centering
    \includegraphics[width=0.44\textwidth]{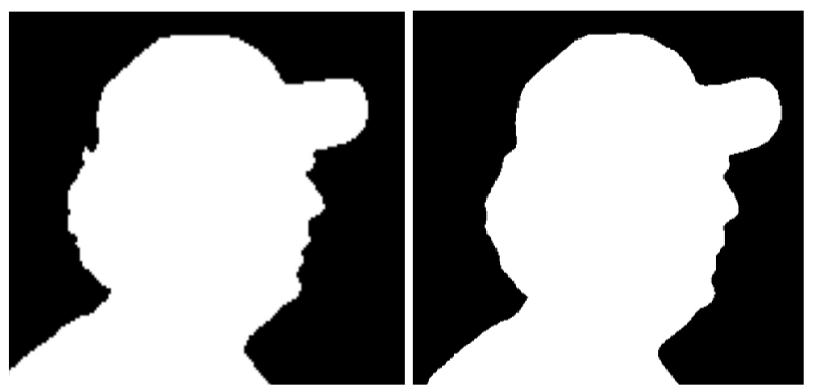}
    \caption{Segmentation boundary quality comparison: (Left) ISEG \cite{Li2018InteractiveIS}, (Right) HyperSeg.}

    \label{fig:mesh1}
\end{figure}

\section{SUMMARY}
The current research presents a novel, high density deep learning algorithm for interactive video segmentation tasks that demonstrates a substantial improvement over current state of the art models with respect to both overall segmentation accuracy and segmentation boundary accuracy. Our research provides several key innovations, including: (1) the application of convolutional tensor decomposition to achieve substantial model compression, (2) the introduction of a novel boundary loss function, (3) a convolutional tessellation technique used to render render pre-trained  features  in  the  native  input  resolution, (4) the application of ”context-aware” skip  connections, and (5) the introduction a new benchmark video segmentation dataset,  the VFX Segmentation Dataset. 

\bibliographystyle{ACM-Reference-Format}

\bibliography{refs.bib}
\end{document}